\documentclass[10pt,twocolumn,letterpaper]{article}

\usepackage{mystyle}
\usepackage{times}
\usepackage{epsfig}
\usepackage{graphicx}
\usepackage{amsmath}
\usepackage{amssymb}
\usepackage{mathrsfs}
\usepackage{caption}
\usepackage{subcaption}

\usepackage[breaklinks=true,bookmarks=false]{hyperref}

\begin{document}

\title{Diversity with Cooperation: Ensemble Methods for Few-Shot Classification} 

\author{Nikita Dvornik,\ \ Cordelia Schmid,\ \ Julien Mairal\\
\\
Univ. Grenoble Alpes, Inria, CNRS, Grenoble INP, LJK, 38000
Grenoble, France\\
{\tt\small firstname.lastname@inria.fr}
}

\maketitle

\def\eg{\emph{e.g.}}
\newcommand\vsp{\vspace*{-0.15cm}}
\newcommand{\pms}[1]{{\scriptsize $\pm$ #1}}

\begin{abstract}
Few-shot classification consists of learning a predictive model that is able to
effectively adapt to a new class, given only a few annotated samples. To solve
this challenging problem, meta-learning has become a popular paradigm that
advocates the ability to ``learn to adapt''. Recent works have shown, however,
that simple learning strategies without meta-learning could be competitive. In
this paper, we go a step further and show that by addressing the fundamental
high-variance issue of few-shot learning classifiers, it is possible to
significantly outperform current meta-learning techniques. Our approach consists
of designing an ensemble of deep networks to leverage the variance of the
classifiers, and introducing new strategies to encourage the networks to
cooperate, while encouraging prediction diversity. Evaluation is conducted on
the mini-ImageNet, tiered-ImageNet and CUB datasets, where we show that even a
single network obtained by distillation yields state-of-the-art results.
\end{abstract}

\section{Introduction}
Convolutional neural networks~\cite{lecun1989backpropagation} have become
standard tools in computer vision to model images, leading to outstanding 
results in many visual recognition tasks such as classification~\cite{krizhevsky2012imagenet},
object detection \cite{blitznet,ssd,faster-rcnn}, or semantic segmentation
\cite{blitznet,long2015fully,ronneberger2015u}. Massively annotated datasets
such as ImageNet~\cite{imagenet} or COCO~\cite{coco} seem to have played a key
role in this success. However, annotating a large corpus is expensive and not
always feasible, depending on the task at hand. Improving the generalization
capabilities of deep neural networks and removing the need for huge sets of annotations
is thus of utmost importance.

While such a grand challenge may be addressed from different
complementary points of views, e.g., large-scale unsupervised
learning~\cite{caron2018deep}, self-supervised
learning~\cite{doersch2017multi,gidaris2018unsupervised}, or by developing regularization
techniques dedicated to deep
networks~\cite{bietti2019kernel,yoshida2017spectral}, we choose in this paper
to focus on variance-reduction principles based on ensemble methods.

Specifically, we are interested in few-shot classification, where a classifier
is first trained from scratch on a medium-sized annotated corpus---that is, without
leveraging external data or a pre-trained network, and then we evaluate its
ability to adapt to new classes, for which only very few annotated samples are
provided (typically 1 or 5). Unfortunately, simply fine-tuning a convolutional
neural network on a new classification task with very few samples has been shown
to provide poor results~\cite{finn2017model}, which has motivated the
community to develop dedicated approaches.

\begin{figure}[t!]
\begin{center}
  \includegraphics[width=0.99\linewidth,trim=25 210 55 20,clip]{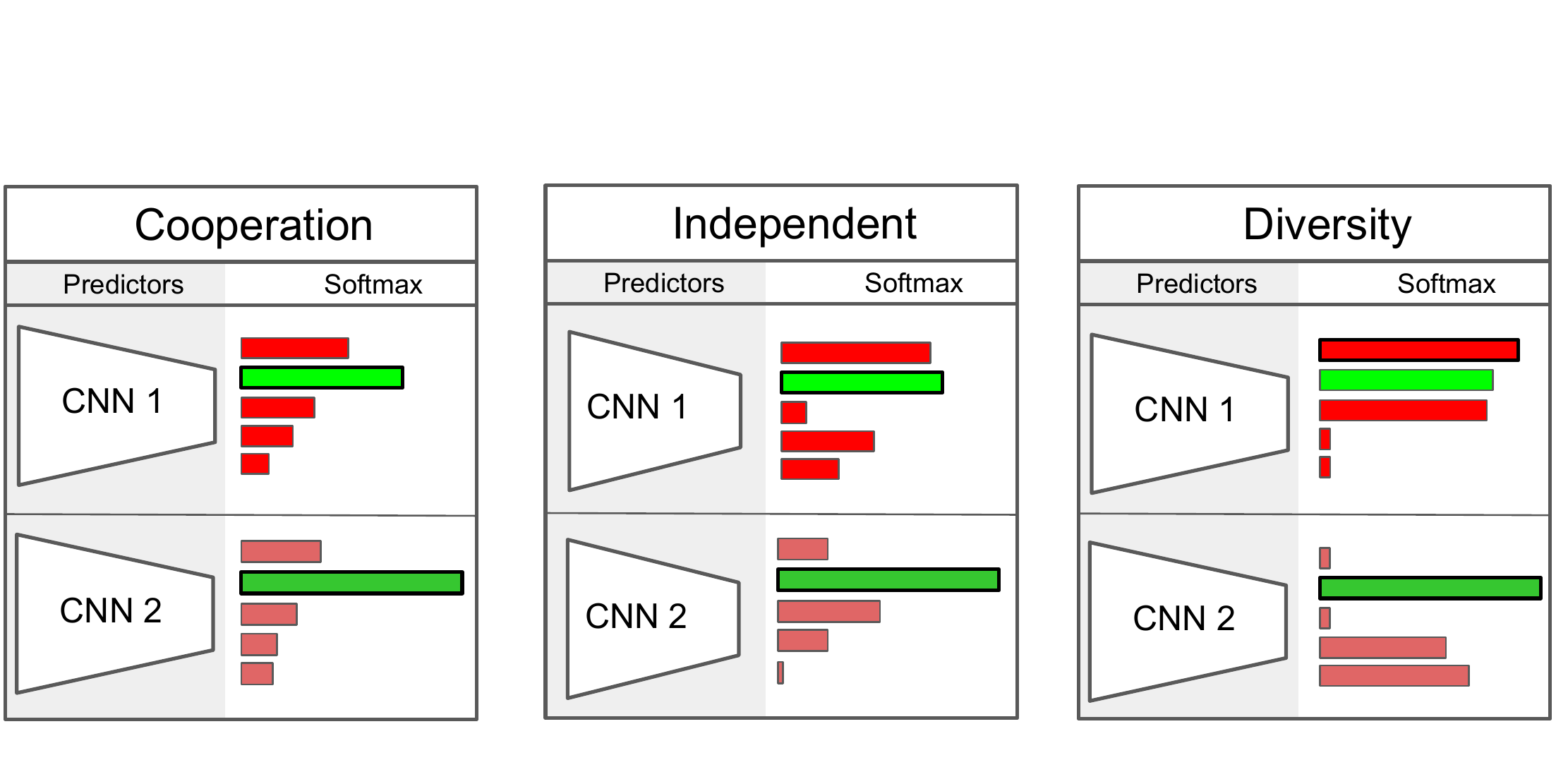}
\end{center}
\vspace*{-0.3cm}
\caption{\textbf{Illustration of the cooperation and diversity strategies on
    two networks.} All networks receive the same image as input and compute
  corresponding class probabilities with softmax. Cooperation encourages the
  non-ground truth probabilities (in red) to be similar, after normalization,
  whereas diversity encourages orthogonality.}
\label{fig:sketch_relationships}
\vspace*{-0.3cm}
\end{figure}

The dominant paradigm in few-shot learning builds upon meta-learning
\cite{finn2017model,ravioptimization,schmidhuber1997shifting,thrun1998lifelong,snell2017prototypical,vinyals2016matching},
which is formulated 
as a principle to learn how to adapt to new learning problems. These approaches
split a large annotated corpus into classification tasks, and the goal is to
transfer knowledge across tasks in order to improve generalization.  While
the meta-learning principle seems appealing for few-shot learning, its empirical benefits
 have not been clearly established
yet. There is indeed strong
evidence~\cite{chen19closerfewshot,gidaris2018dynamic,qiao2018few} that
training CNNs from scratch using meta-learning performs substantially worse
than if CNN features are trained in a standard fashion---that is, by minimizing
a classical loss function relying on corpus annotations; on the other hand, learning only
the last layer with meta-learning has been found to produce better
results~\cite{gidaris2018dynamic,qiao2018few}. Then, it was recently shown
in~\cite{chen19closerfewshot} that simple distance-based classifiers could
achieve similar accuracy as meta-learning approaches.

Our paper goes a step further and shows that 
meta-learning-free approaches can be improved and significantly outperform the
current state of the art in few-shot learning. Our angle of attack consists of using ensemble methods to reduce the variance
of few-shot learning classifiers, which is inevitably high given the small
number of annotations.  Given an initial
medium-sized dataset (following the standard setting of few-shot learning), 
the most basic ensemble approach consists of first training several CNNs independently before freezing them and removing the last prediction layer. Then, given
a new class (with few annotated samples), we build a mean centroid
classifier for each network and estimate class probabilities---according to a
basic probabilistic model---of test samples based on the distance to the
centroids~\cite{mensink2013distance,snell2017prototypical}. The obtained
probabilities are then averaged over networks, resulting in 
higher accuracy. 

While we show that the basic ensemble method where networks are trained independently already performs well, we
introduce penalty terms that allow the networks to cooperate
during training, while encouraging enough diversity of predictions, as illustrated in Figure~\ref{fig:sketch_relationships}. The
motivation for cooperation is that of easier learning and regularization, where individual
networks from the ensemble can benefit from each other. The motivation
for encouraging diversity is classical for ensemble methods~\cite{dietterich2000ensemble},
where a collection of weak learners making diverse predictions often performs
better together than a single strong one. Whereas these two principles seem
in contradiction with each other at first sight, we show that both principles
are in fact useful and lead to significantly better results than the basic ensemble method. 
Finally, we also show that a single network trained by
distillation~\cite{hinton2015distilling} to mimic the behavior of the ensemble
also performs well, which brings a significant speed-up at test time.
In summary, our contributions are three-fold:
\begin{itemize}
\item We introduce mechanisms to encourage cooperation and diversity  for learning an ensemble of networks. We study these two
principles for few-shot learning and characterize the regimes where they are useful.
\item We show that it is possible to significantly outperform current
state-of-the-art techniques for few-shot classification without using
meta-learning.
\item As a minor contribution, we also show how to distill an ensemble into a single
network with minor loss in accuracy, by using additional unlabeled data.
\end{itemize}

\section{Related Work}\label{sec:related}
In this section, we discuss related work on few-shot learning, meta-learning,
and ensemble methods.

\vsp
\paragraph{Few-shot classification.} Typical few-shot classification
problems consist of two parts called meta-training and
meta-testing~\cite{chen19closerfewshot}. During the meta-training stage, one
is given a large-enough annotated dataset, which is used to train a predictive model.  During meta-testing, novel categories
are provided along with few annotated examples, and we evaluate the capacity of the predictive model to 
retrain or adapt, and then generalize on these new classes.

Meta-learning approaches typically sample few-shot learning classification tasks from the meta-training
dataset, and train a model such that it should generalize on a new
task that has been left aside. For instance, in \cite{finn2017model}
a ``good network initialization'' is learned such that a small number of
gradient steps on a new problem is sufficient to obtain a good solution.
In~\cite{ravioptimization}, the authors learn both the network
initialization and an update rule (optimization model) represented by a
Long-Term-Short-Memory network (LSTM).  Inspired by few-shot learning
strategies developed before deep learning approaches became
popular~\cite{mensink2013distance}, distance-based classifiers based on the
distance to a centroid were also proposed, \eg, prototypical
networks~\cite{snell2017prototypical}, or more sophisticated classifiers with
attention~\cite{vinyals2016matching}. All these methods consider a classical
backbone network, and train it from scratch using meta-learning.

Recently, these meta-learning were found to be sub-optimal \cite{gidaris2018dynamic,oreshkin2018tadam,qiao2018few}. Specifically, better results were obtained by 
training the network on the classical classification task using the meta-training
data in a first step, and then only fine-tuning with meta-learning in a second
step~\cite{oreshkin2018tadam,rusu2018meta,ye2018learning}.
Others such as~\cite{gidaris2018dynamic,qiao2018few} simply freeze the network obtained
in the first step, and train a simple prediction layer with meta-learning,
which results in similar performance. Finally, the
paper~\cite{chen19closerfewshot} demonstrates
that simple baselines without meta-learning---based on distance-based
classifiers---work equally well. 
Our paper pushes such principles even further and shows that by appropriate variance-reduction
techniques, these approaches can significantly outperform the current state of the art.

\vsp
\paragraph{Ensemble methods.} It is well known that ensemble methods reduce the
variance of estimators and subsequently may improve the quality of
prediction~\cite{friedman2001elements}. To gain accuracy from averaging,
various randomization or data augmentation techniques are typically used to
encourage a high diversity of
predictions~\cite{breiman1996heuristics,dietterich2000ensemble}.  While
individual classifiers of the ensemble may perform poorly, the quality of the
average prediction turns out to be sometimes surprisingly high.

Even though ensemble methods are costly at training time for neural
networks, it was shown that a single network trained to
mimic the behavior of the ensemble could perform almost equally
well~\cite{hinton2015distilling}---a procedure called distillation---thus
removing the overhead at test time. To improve the scalability of distillation 
in the context of highly-parallelized implementations, an online distillation
procedure is proposed in~\cite{anil2018large}. There, each network is
encouraged to agree with the averaged predictions made by other networks of the
ensemble, which results in more stable models.
The objective  of our work is however significantly different. The
form of cooperation they encourage between networks is indeed targeted to
scalability and stability (due to industrial constraints), but online distilled
networks do not necessarily perform better than the basic ensemble strategy.
Our goal, on the other hand, is to improve the quality of prediction and do
better than basic ensembles.

To this end, we encourage cooperation in a different manner, by encouraging
predictions between networks to match in terms of class probabilities
conditioned on the prediction not being the ground truth label.
While we show that such a strategy alone is useful in general when the number
of networks is small, encouraging diversity becomes crucial when this number grows.
Finally, we show that distillation can help to reduce the computational overhead at test time.

\section{Our Approach}\label{sec:approach}
\newcommand{\m}{\mathcal}
In this section, we present our approach for few-shot classification,
starting with preliminary components.
\subsection{Mean-centroid classifiers}
We now explain how to perform few-shot classification with a fixed feature
extractor and a mean centroid classifier.

\vsp
\paragraph{Few-shot classification with prototype classifier.}\label{subsec:prot}
During the meta-training stage, we are given a dataset
$D_b$ with annotations, which we use to train a prediction function $f_{\theta}$ represented by a
CNN. Formally, after training the CNN on~$D_b$, we remove the final prediction
layer and use the resulting vector $\tilde{f}_\theta(x)$ as a set of visual features
for a given image~$x$. The parameters $\theta$ represent the weights of the
network, which are frozen after this training step.

During meta-testing, we are given a new
dataset $D_q = \{x_i, y_i\}_{i=1}^{nk}$, where $n$ is a number of new
categories and $k$ is the number of available examples for each class. The $(x_i,y_i)$'s represent image-label pairs. Then, we build a mean centroid classifier, leading to the class prototypes
\begin{equation}
   c_j = \frac{1}{k} \sum_{i=1}^{k}{\tilde{f}_\theta(x_i)}, \qquad j=1,...,n.~\label{eq:mean}
\end{equation}
Finally, a test sample $x$ is assigned to the nearest centroid's class. Simple mean-centroid classifiers have proven to be effective in
the context of few-shot classification \cite{chen19closerfewshot,mensink2013distance,snell2017prototypical},
which is confirmed in the following experiment.

\vsp
\paragraph{Motivation for mean-centroid classifier.}
We report here an experiment showing that a more complex model than~(\ref{eq:mean})
does not necessarily lead to better results for few-shot learning.
Consider indeed a parametrized version of~(\ref{eq:mean}):
\begin{equation}
   c_j = \sum_{i=1}^{nk}{\alpha_i^j \tilde{f}_\theta(x_i)}, \qquad j=1,...,n, \label{eq:mean2}
\end{equation}
where the weights $\alpha_i^j$ can be learned with gradient descent by
maximizing the likelihood of the probabilistic model
\begin{equation}\label{eq:probs}
   p_j(y = l | x) = \frac{\exp(-d(\tilde{f}_\theta(x), c_l))}{\sum_{j=1}^n{\exp(-d(\tilde{f}_\theta(x), c_j)}}
\end{equation}
where $d(\cdot, \cdot)$ is a distance function, such as Euclidian distance or
negative cosine similarity. Since the coefficients are learned from data and
not set arbitrarily to $1/k$ as in~(\ref{eq:mean}), one would potentially expect this
method to produce better classifiers if appropriately regularized.
When we run the evaluation of the aforementioned classifiers on 1000 5-shot
learning tasks sampled from \texttt{miniImagenet-test} (see experimental section for details about this dataset), we get similar
results on average: $77.28 \pm 0.46 \%$ for~(\ref{eq:mean}) vs.
$77.01 \pm 0.50\%$ for~(\ref{eq:mean2}), confirming that learning meaningful parameters in this
very-low-sample regime is difficult.

%

\subsection{Learning ensembles of deep networks} 
During meta-training, one needs to minimize the following loss
function over a training set $\{x_i, y_i\}_{i=1}^{m}$:
\begin{equation}\label{eq:single_loss}
  L(\theta) = \frac{1}{m} \sum_{i=1}^m{\ell(y_i, \sigma(f_{\theta}(x_i)))} + \lambda \|\theta\|^2_{2},
\end{equation}
where $f_{\theta}$ is a CNN as before.
The cost function $\ell(\cdot, \cdot)$ is the cross-entropy between
ground-truth labels and predicted class probabilities $p=\sigma
(f_{\theta}(x))$, where $\sigma$ is the normalized
exponential function, and $\lambda$ is a weight decay parameter.

When training an ensemble of $K$ networks $f_{\theta_k}$ independently, one
would solve~(\ref{eq:single_loss}) for each network separately.
While these terms may look identical, solutions provided by deep neural
networks will typically differ when trained with different initializations and random seeds,
making ensemble methods appealing in this context.

In this paper, we are interested in ensemble of networks, but we also want to model
relationships between its members; this may be achieved by considering a pairwise
penalty function $\psi$, leading to the joint formulation:

\begin{multline}\label{eq:total_loss}
  L(\bar{\theta}) =  \sum_{j=1}^K \left(\frac{1}{n} \sum_{i=1}^n {\ell(y_i, \sigma(f_{\theta_j}(x_i)))} + \lambda \|\theta_j\|^2_{2}\right)\\ + \frac{\gamma}{n (K-1)}\sum_{i=1}^n \sum_{\substack{j,l \\ j\neq l} }^K \psi (y_i, f_{\theta_j}(x_i),f_{\theta_l}(x_i)),
\end{multline}
where $\bar{\theta}$ is the vector obtained by concatenating all the parameters $\theta_j$.
By carefully designing the function~$\psi$ and setting up appropriately the
parameter  $\gamma$, it is possible to achieve desirable properties of the
ensemble, such as diversity of predictions or collaboration during training.

\begin{figure*}[t!]
\begin{center}
\begin{subfigure}{.45\textwidth}
    \includegraphics[width=0.99\textwidth,trim=20 1 30 30,clip]{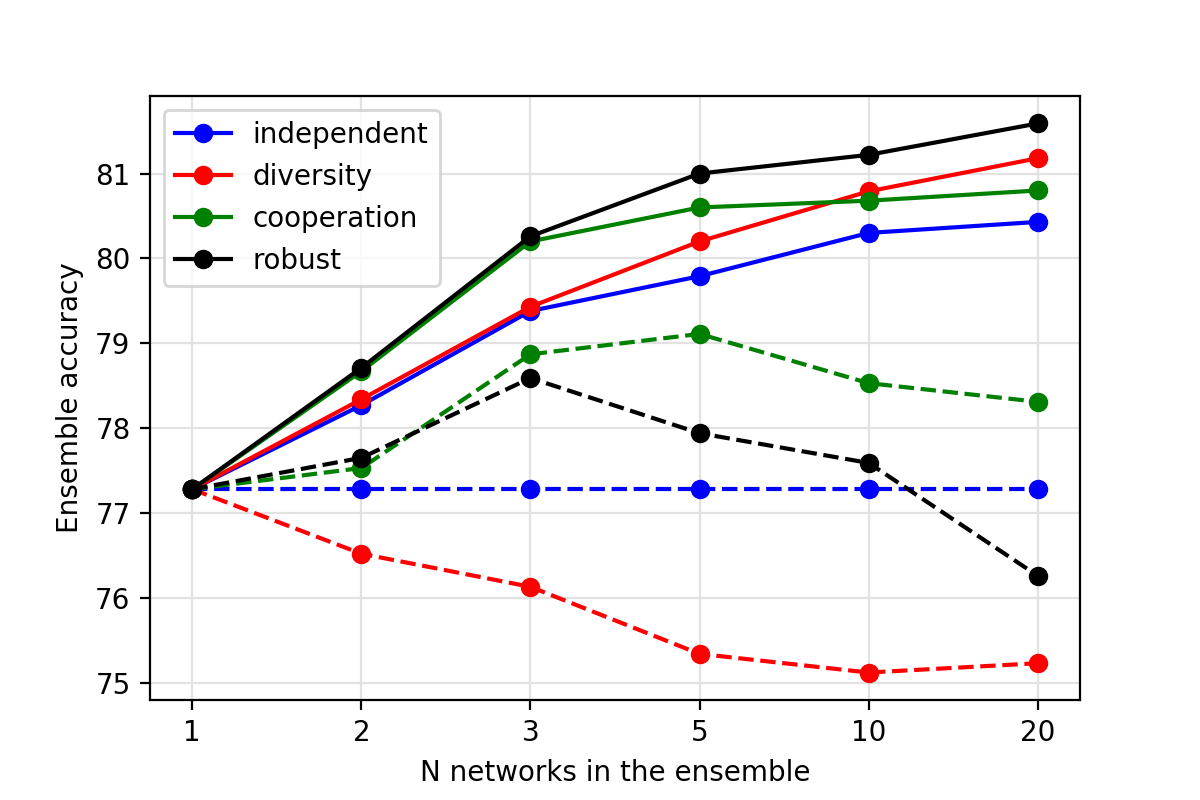}
  \caption{\textit{Mini}ImageNet 5-shots}
\end{subfigure}
\begin{subfigure}{.45\textwidth}
    \includegraphics[width=0.99\textwidth,trim=20 1 30 30,clip]{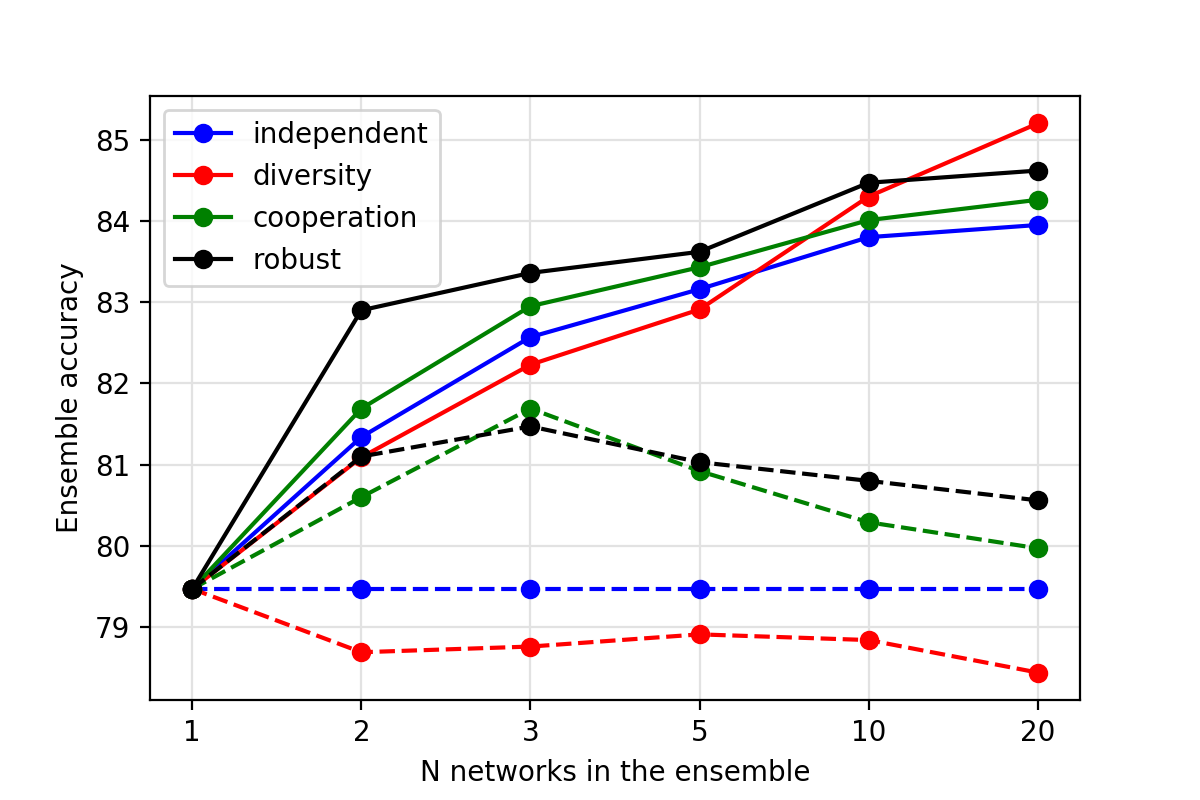}
  \caption{CUB 5-shots}
\end{subfigure}
\end{center}
\vspace*{-0.4cm}
\caption{\textbf{Accuracies of different ensemble strategies (one for each color) for various numbers
    of networks.}
  Solid lines give the ensemble accuracy after aggregating predictions. The
  average performance of single models from the ensemble is plotted with a
  dashed line. 
Best viewed in color.}
\label{fig:ens_perf}
\vspace*{-0.3cm}
\end{figure*}

\subsection{Encouraging diversity and cooperation}\label{sec:emb_ens}
To reduce the high variance of few-shot learning classifiers, we use
ensemble methods trained with a particular interaction function~$\psi$, as
in~(\ref{eq:total_loss}).
Then, once the parameters~$\theta_j$ have been learned during meta-training,
classification in meta-testing is performed by considering a collection of $K$
mean-centroid classifiers associated to the basic probabilistic model presented in Eq.~(\ref{eq:probs}).
Given a test image, the $K$ class probabilities are averaged. Such a strategy
was found to perform empirically better than a voting scheme. 

As we show in the experimental section, the choice of pairwise relationship
function~$\psi$ significantly influences the quality of the ensemble.
Here, we describe three different strategies, which all provide benefits in
different regimes, starting by a criterion encouraging diversity of predictions.

\vsp
\paragraph{Diversity.} 
One way to encourage diversity consists of introducing
randomization in the learning procedure, \eg, by using data augmentation~\cite{breiman1996heuristics,friedman2001elements} or
various initializations. Here, we also evaluate the effect of an interaction 
function~$\psi$ that acts directly on the network predictions.
Given an image $x$, two models parametrized by~$\theta_i$ and~$\theta_j$ respectively lead
to class probabilities $p_i=\sigma(f_{\theta_i}(x))$ and
$p_j=\sigma(f_{\theta_j}(x))$.  During training, $p_i$ and $p_j$
are naturally encouraged to be close to the assignment vector $e_y$ in
$\{0,1\}^d$ with a single non-zero entry at position $y$, where $y$ is the
class label associated to $x$ and $d$ is the number of classes.

From~\cite{hinton2015distilling}, we know that even though only the
largest entry of $p_i$ or $p_j$ is used to make predictions, other entries---typically not corresponding to the ground truth label $y$---carry important
information about the network.  It becomes then natural to consider the
probabilities $\hat{p}_i$ and $\hat{p}_j$ conditioned on not being the ground
truth label. Formally, these are obtained by setting to zero the entry $y$ in
$p_i$ and $p_j$ renormalizing the corresponding vectors such that they sum to one.
Then, we consider the following diversity penalty
\begin{equation}\label{eq:diversity}
  \phi(\hat{p}_i, \hat{p}_j) = \text{cos}(\hat{p}_i, \hat{p}_j).
\end{equation}
When combined with the
loss function, the resulting formulation encourages the networks to make the right
prediction according to the ground-truth label, but then they are also encouraged to make different
second-best, third-best, and so on, choice predictions (see Figure~\ref{fig:sketch_relationships}).
This penalty turns out to be particularly effective when the number of networks is large,
as shown in the experimental section. It typically worsens the performance of 
individual classifiers on average, but make the ensemble prediction more accurate.

\paragraph{Cooperation.}
Apparently opposite to the previous principle, encouraging the conditional
probabilities $\hat{p}_i$ to be similar---though with a different metric---may
also improve the quality of prediction by allowing the networks
to cooperate for better learning. Our experiments show that such a principle
alone may be effective, but it appears to be mostly useful when the number of
training networks is small, which suggests that there is a trade-off
between cooperation and diversity that needs to be found.

Specifically, our experiments show that using the negative
cosine---in other words, the opposite of~(\ref{eq:diversity})---is ineffective.
However, a penalty such as the symmetrized KL-divergence turned out to 
provide the desired effect:
\begin{equation}\label{eq:cooperation}
  \phi(\hat{p}_i, \hat{p}_j) = \frac{1}{2} (\text{KL}(\hat{p}_i || \hat{p}_j) + \text{KL}(\hat{p}_j || \hat{p}_i)).
\end{equation}
By using this penalty, we managed to obtain more stable and faster training,
resulting in better performing individual networks,
but also---perhaps surprisingly---a better ensemble.
Unfortunately, we also observed that the gain of ensembling diminishes with the
number of networks in the ensemble since the individual members become too
similar.

\vsp
\paragraph{Robustness and cooperation.}
Given experiments conducted with the two previous penalties, a trade-off
between cooperation and diversity seems to correspond to two regimes (low vs.
high number of networks). This motivated us to develop an approach designed
to achieve the best trade-off. When considering the cooperation penalty~(\ref{eq:cooperation}),
we try to increase diversity of prediction by several additional means.
i) We randomly drop some networks from the ensemble at each
training iteration, which causes the networks to learn on different data streams
and reduces the speed of knowledge propagation. ii) We introduce Dropout
within each network to increase randomization. iii) We feed each network with a different (crop, color)
transformation of the same image, which makes the ensemble more
robust to input image transformations. Overall, this strategy was found to perform best
in most scenarios (see Figure~\ref{fig:ens_perf}).

\subsection{Ensemble distillation}
As most ensemble methods, our ensemble strategy introduces a significant
computional overhead at training time. To remove the overhead at test time,
we use a variant of knowledge distillation~\cite{hinton2015distilling}
to compress the ensemble into a single network~$f_w$.
Given the meta-training dataset $D_b$, we consider the following 
cost function on example~$(x,y)$:
\begin{multline}
  \ell(x, y) = (1 - \alpha) \cdot \hat{\ell}(e_y, \sigma(f_{w}(x))) \\
  - \alpha \cdot T^2 \cdot \hat{\ell} \left(\tfrac{1}{K} \textstyle \sum_{k=1}^K \sigma \left(\tfrac{f_{\theta_k}(x)}{T}\right), \ \sigma \left(\tfrac{f_{w}(x)}{T}\right)\right), \label{eq:distil}
\end{multline}
where, $\hat{\ell}$ is cross-entropy, $e_y$ is a one-hot embedding of the
true label $y$. The second term performs distillation with parameter $T$ (see~\cite{hinton2015distilling}). It encourages the single
model~$f_w$ to be similar to the average output of the ensemble. In our experiments,
we are able to obtain a model with performance relatively close to that of the
ensemble (see Section~\ref{sec:exp}).

\vsp
\paragraph{Modeling out-of-distribution behavior.} When distillation is performed on
the dataset $D_b$, the network~$f_w$ mimics the behavior of the
ensemble on a specific data distribution. However, new categories are
introduced at test time. Therefore, we also tried distillation by using
additional unnannotated data, which yields slightly better performance.
%

\section{Experiments}\label{sec:exp}
We now present experiments to study the effect of cooperation
and diversity for ensemble methods, and start with 
experimental and implementation details.

\subsection{Experimental Setup}\label{sec:setup}
\paragraph{Datasets.} We use 
\textit{mini}-ImageNet~\cite{ravioptimization} and
\textit{tiered}-ImageNet~\cite{ren2018meta} which are derived from the original
ImageNet~\cite{imagenet} dataset and Caltech-UCSD Birds (CUB)
200-2011~\cite{WahCUB_200_2011}. \textit{Mini}-ImageNet consists of 100
categories---64 for training, 16 for validation and 20 for testing---with 600
images each. \textit{Tiered}-ImageNet is also a subset of ImageNet that includes
351 class for training, 97 for validation and 160 for testing which is 779,165
images in total. The splits are chosen such that the training classes are
sufficiently different from the test ones, unlike in \textit{mini}-ImageNet. The
CUB dataset consists of 11,788 images of birds of more than 200 species. We
adopt train, val, and test splits from~\cite{ye2018learning}, which were
originally created by randomly splitting all 200 species in 100 for training, 50
for validation, and 50 for testing.

\vsp
\paragraph{Evaluation.} In few-shot classification, the test set is used to sample
$N$ 5-way classification problems, where only $k$ examples of each category are
provided for training and 15 for evaluation. We
follow~\cite{finn2017model,gidaris2018dynamic,oreshkin2018tadam,qiao2018few,ravioptimization}
and test our algorithms for $k =$ 1 and 5 and $N$ is set to $1\,000$. Each time, classes
and corresponding train/test examples are sampled at random. For all our
experiments we report the mean accuracy (in \%) over $1\,000$ tasks and $95 \%$
confidence interval.

\vsp
\paragraph{Implementation details.} For all experiments, we use the Adam
optimizer~\cite{adam} with an initial learning rate $10^{-4}$, which is
decreased by a factor 10 once during training when no improvement in validation
accuracy is observed for $p$ consecutive epochs. For {\it mini}-ImageNet, we use
$p=10$, and 20 for the CUB dataset. When distilling an ensemble into one
network, $p$ is doubled. We use random crops and color augmentation during
training as well as weight decay with parameter~$\lambda=$ $5 \cdot 10^{-4}$.
All experiments are conducted with the ResNet18 architecture~\cite{resnet},
which allows us to train our ensembles of $20$ networks on a single GPU. Input
images are then re-scaled to the size $224 \times 224$, and organized in
mini-batches of size 16. Validation accuracy is computed by running 5-shot
evaluation on the validation set. During the meta-testing stage, we take central
crops of size $224 \times 224$ from images and feed them to the feature
extractor. No other preprocessing is used at test time. When building a mean
centroid classifier, the distance $d$ in~(\ref{eq:probs}) is computed as the
negative cosine similarity~\cite{snell2017prototypical}, which is rescaled by a
factor 10. \\
For a fair comparison, we have also evaluated ensembles
composed of ResNet18~\cite{resnet} with input image size $84 \times 84$ and
WideResNet28~\cite{Zagoruyko2016WRN} with input size $80 \times 80$. All details
are reported in Appendix. For reproducibility purposes, our implementation will be made
available at \url{http://thoth.inrialpes.fr/research/fewshot_ensemble/}.

\subsection{Ensembles for Few-Shot Classification}\label{sec:ensembles_exp}
In this section, we study the effect of ensemble training with pairwise interaction
terms that encourage cooperation or diversity. For that purpose, we analyze the link
between the size of
ensembles and their 1- and 5-shot classification performance on the
\textit{mini}-ImageNet and CUB datasets.

\vsp
\paragraph{Details about the three strategies.}
When models are trained jointly, the data stream is shared across all 
networks and weight updates happen simultaneously. This is achieved by placing
all models on the same GPU and optimizing the loss~(\ref{eq:total_loss}).  When
training a diverse ensemble, we use the cosine function~(\ref{eq:diversity})
and selected the parameter $\gamma=1$ that performed best on the validation set among the tested values
($10^i$, for $i=-2,\ldots,2$) for $n=5$ and $n=10$ 
networks. Then, this value was kept for other values of $n$. To enforce
cooperation between networks, we use the symmetrized KL
function~(\ref{eq:cooperation}) and selected the parameter $\gamma=10$ in the
same manner. 
Finally, the robust ensemble strategy is trained with the cooperation relationship penalty and the
same parameter $\gamma$, but we use
Dropout with probability 0.1 before the last layer;
each of the network is dropped from the ensemble with probability 0.2 at 
every iteration; different networks receive different transformation of the
same image, i.e. different random crops and color augmentation. 

\vsp
\paragraph{Results.}
Table~\ref{tab:manyens_mini} and Table A1 of Appendix
summarize the few-shot classification accuracies of ensembles trained with our strategies and compare with basic ensembles. On the \textit{mini}-ImageNet dataset,
the results for 1- and 5-shot classification are consistent with each other. Training with cooperation allows smaller ensembles ($n\le5$) to
perform better, which leads to higher individual accuracy of the
ensemble members, as seen in Figure~\ref{fig:ens_perf}. However, when $n \geq
10$, cooperation is less effective, as opposed to the diversity strategy, which
benefits from larger $n$. As we can see from Figure~\ref{fig:ens_perf},
individual members of the ensemble become worse, but the ensemble accuracy
improves substantially. Finally, the robust strategy seems to perform best for
all values of $n$ in almost all settings. The situation for the CUB dataset is
similar, although we notice that robust ensembles perform similarly as the
diversity strategy for $n=20$.

\begin{table*}[t!] 
\centering
\small\addtolength{\tabcolsep}{-10pt}
\renewcommand{\arraystretch}{1.0}
\renewcommand{\tabcolsep}{1.6mm}
\resizebox{0.9\linewidth}{!}{
\begin{tabular}{l| c c c c c c}
\multicolumn{7}{c}{\bfseries 5-shot}\\ \hline
  Ensemble type & 1 & 2 & 3 & 5 & 10 & 20\\
  \hline
  Independent & 77.28 \pms{0.46}& 78.27 \pms{0.45} & 79.38 \pms{0.43} & 80.02 \pms{0.43}&  80.30 \pms{0.43} & 80.57 \pms{0.42} \\

  Diversity   & 77.28 \pms{0.46}& 78.34 \pms{0.46} & 79.18 \pms{0.43} & 79.89 \pms{0.43}&  80.82 \pms{0.42} & 81.18 \pms{0.42} \\

  Cooperation & 77.28 \pms{0.46}& 78.67 \pms{0.46} & 80.20 \pms{0.42} & 80.60 \pms{0.43}&  80.72 \pms{0.42} & 80.80 \pms{0.42} \\
  \hline                                        
  Robust      & 77.28 \pms{0.46}& 78.71 \pms{0.45} & 80.26 \pms{0.43} & 81.00 \pms{0.42}&  81.22 \pms{0.43} & 81.59 \pms{0.42} \\
  \hline
  \multicolumn{7}{l}{Distilled Ensembles}\\
  \hline
  Robust-\textit{dist}     & $-$ & 79.44 \pms{0.44} & 79.84 \pms{0.44} & 80.01 \pms{0.42}&  80.25 \pms{0.44} & 80.63 \pms{0.42} \\
  Robust-\textit{dist}++  & $-$ & 79.16 \pms{0.46} & 80.00 \pms{0.44} & 80.25 \pms{0.42}&  80.35 \pms{0.44} & 81.19 \pms{0.43} \\
  \hline
\end{tabular}
}
\vspace{0.4cm}
\resizebox{0.9\linewidth}{!}{
\begin{tabular}{l| c c c c c c}
  \multicolumn{7}{c}{\bfseries 1-shot}\\ \hline
  Ensemble type & 1 & 2 & 3 & 5 & 10 & 20 \\
  \hline
  Independent & 58.71 \pms{0.62} & 60.04 \pms{0.60} & 60.83 \pms{0.63} & 61.34 \pms{0.61} & 61.93 \pms{0.61} & 62.06 \pms{0.61} \\

  Diversity   & 58.71 \pms{0.63} & 59.95 \pms{0.61} & 61.27 \pms{0.62} & 61.43 \pms{0.61} & 62.23 \pms{0.61} & 62.47 \pms{0.62} \\

  Cooperation & 58.71 \pms{0.62} & 60.20 \pms{0.61} & 61.46 \pms{0.61} & 61.61 \pms{0.61} & 62.06 \pms{0.61} & 62.12 \pms{0.62} \\
  Robust      & 58.71 \pms{0.62} & 60.91 \pms{0.62} & 62.36 \pms{0.60} & 62.70 \pms{0.61} & 62.97 \pms{0.62} & 63.95 \pms{0.61} \\
  \hline
  \multicolumn{7}{l}{Distilled Ensembles}\\
  \hline
  Robust-\textit{dist}     & $-$ & 62.33 \pms{0.62} & 62.64 \pms{0.60} & 63.14 \pms{0.61} & 63.01 \pms{0.62} & 63.06 \pms{0.61} \\
  Robust-\textit{dist} ++  & $-$ & 62.07 \pms{0.62} & 62.81 \pms{0.60} & 63.39 \pms{0.61} & 63.20 \pms{0.62} & 63.73 \pms{0.62} \\
  \hline
\end{tabular}
}
\vspace{0.1cm}
\caption{\textbf{Few-shot classification accuracy on \textit{mini}-ImageNet.}
  The first column gives the strategy, the top row indicates the number~$N$
  of networks in an ensemble. Here, \textit{dist} means that an ensemble was
  distilled into a single network, and '++' indicates that extra unannotated
  images were used for distillation. We performed 1\,000 independent experiments on
  {\it mini}-ImageNet-test and report the average with 95\% confidence interval.
All networks are trained on {\it mini}-ImageNet-train set.}
\label{tab:manyens_mini}
\end{table*}

\subsection{Distilling an ensemble}\label{sec:distill_exp}
We distill robust ensembles of all sizes to study knowledge transferability
with growing ensemble size. To do so, we use the meta-training dataset and optimize
the loss~(\ref{eq:distil}) with parameters $T=10$ and $\alpha=0.8$. 
For the strategy using external data, we randomly add at each iteration 8 images (without annotations)
from the COCO~\cite{coco} dataset to the 16 annotated samples from the meta-training
data. Those images contribute only to the distillation part of
the loss~(\ref{eq:distil}).
Table \ref{tab:manyens_mini} and Table A1 of Appendix display model
accuracies for {\it mini}-ImageNet and CUB datasets respectively. For 5-shot
classification on {\it mini}-ImageNet, the difference between ensemble
 and its distilled version is rather low (around 1\%), while adding extra
non-annotated data helps reducing this gap. Surprisingly, 1-shot classification
accuracy is slightly higher for distilled models than for their corresponding full
ensembles. 
On the CUB dataset, distilled models stop improving after $n=5$, even though the
performance of full ensembles keeps growing. This seems to indicate that the 
capacity of the single network may have been reached, which suggests using a more
complex architecture here. Consistently with such hypothesis, adding extra data is not as helpful as for {\it
mini}-ImageNet, most likely because data distributions of COCO and CUB are more
different.

In Tables~\ref{tab:comp_to_base},~\ref{tab:tiered}, we also compare the performance of our
distilled networks with other baselines from the literature, including current
state-of-the-art meta-learning approaches, showing that our approach does
significantly better on the {\it mini}-ImageNet~\cite{ravioptimization} and {\it
tiered}-ImageNet~\cite{ren2018meta} datasets.


\begin{table}[t!] 
\centering
\renewcommand{\arraystretch}{1.0}
\renewcommand{\tabcolsep}{1.6mm}
\resizebox{0.99\linewidth}{!}{
\begin{tabular}{l| c c c c}
  Method & Input size & Network & 5-shot & 1-shot \\
  \hline
  TADAM \cite{oreshkin2018tadam}               & 84  & ResNet    & 76.70 \pms{0.30} & 58.50 \pms{0.30} \\
  Cosine + Attention \cite{gidaris2018dynamic} & 224 & ResNet    & 73.00 \pms{0.64} & 56.20 \pms{0.86} \\
  Linear Classifier \cite{chen19closerfewshot} & 224 & ResNet    & 74.27 \pms{0.63} & 51.75 \pms{0.80} \\
  Cosine Classifier \cite{chen19closerfewshot} & 224 & ResNet    & 75.68 \pms{0.63} & 51.87 \pms{0.77} \\
  PPA \cite{qiao2018few}                       & 80 & WideResNet & 73.74 \pms{0.19} & 59.60 \pms{0.41} \\
  LEO \cite{rusu2018meta}                      & 80 & WideResNet & 77.59 \pms{0.12} & 61.76 \pms{0.08} \\
  FEAT \cite{ye2018learning}                   & 80 & WideResNet & 78.32 \pms{0.16} & 61.72 \pms{0.11} \\
  \hline
  Robust 20-\textit{dist}++ (ours)             & 224 & ResNet     & {\bf 81.19} \pms{0.43} & {\bf 63.73} \pms{0.62} \\
  Robust 20-\textit{dist}++ (ours)             & 84 & ResNet      & 75.62 \pms{0.48} & 59.48 \pms{0.62} \\
  Robust 20-\textit{dist}++ (ours)             & 80 & WideResNet  & 81.17 \pms{0.43} & 63.28 \pms{0.62} \\
  \hline
  Robust 20 Full                               & 224  & ResNet      & 81.59 \pms{0.42} & 63.95 \pms{0.61} \\
  Robust 20 Full                               & 84   & ResNet      & 76.90 \pms{0.42} & 59.38 \pms{0.65} \\
  Robust 20 Full                               & 80   & WideResNet  & 81.94 \pms{0.44} & 63.46 \pms{0.62} \\
  \hline
\end{tabular}
}
\vspace{0.3cm}
\caption{\textbf{Comparison of distilled ensembles to other methods on 1- and
    5-shot \textit{mini}ImageNet.}
  The two last columns
  display the accuracy on 1- and 5-shot learning tasks.  To evaluate our methods we performed $1\,000$
  independent experiments on \textit{Mini}ImageNet-test and report the average and
  $95\%$ confidence interval. Here, '++' means that extra non-annotated images
  were used to perform distillation. The last model is a full ensemble and should
  not be directly compared to the rest of the table.}
\label{tab:comp_to_base}
\end{table}

\subsection{Study of relationship penalties}\label{sec:design}
There are many possible ways to model relationship between the members of an
ensemble. In this subsection, we study and discuss such particular choices.

\paragraph{Input to relationship function.} 
As noted by~\cite{hinton2015distilling}, class probabilities obtained by the
softmax layer of a network seem to carry a lot of information and are useful for distillation. However,
after meta-training, such probabilities are often close to binary vectors
with a dominant value associated to the ground-truth label.
To make small values more noticeable, distillation uses a parameter~$T$, as
in~(\ref{eq:distil}). Given such a class probability computed by a network, we
experimented such a strategy consisting of introducing new probabilities $\hat{p}=\sigma(p/T)$,
where the contributions of non ground-truth values are emphasized.
When used within our diversity~(\ref{eq:diversity}) or
cooperation~(\ref{eq:cooperation}) penalties, we however did not see any 
improvement over the basic ensemble method.
Instead, we found that computing the class probabilities conditioned on not being the ground truth label, as explained in Section~\ref{sec:emb_ens}, would perform much better.

This is illustrated on the following experiment with two network ensembles of size $n=5$.
We enforce similarity on the full probability vectors in the first one, computed
with softmax at $T=10$ following~\cite{anil2018large}, and with
conditionally non-ground-truth
probabilities for the second one as defined in Section~\ref{sec:emb_ens}. When using the cooperation training formulation,
the second strategy turns out to perform about  1\% better than the first one
(79.79 \% vs 80.60\%), when tested on \textit{Mini}ImageNet. Similar observations
have been made using the diversity criterion. In comparison, the basic ensemble method without interactions
achieves about $80\%$.

\paragraph{Choice of relationship function.} 
In principle, any similarity measure could be used to design a penalty 
encouraging cooperation. 
%
Here, we show that in fact, selecting the right criterion for comparing
probability vectors (cosine similarity, L2 distance, symmetrized KL
divergence), is crucial depending on the desired effect (cooperation or
diversity).  In Table~\ref{tab:criteria_choice}, we perform such a comparison
for an ensemble with $n=5$ networks on the \textit{Mini}ImageNet dataset for a
$5-$shot classification task, when plugging the above function in the
formulation~\ref{eq:total_loss}, with a specific sign.  The parameter $\gamma$
for each experiment is chosen such that the performance on the validation set
is maximized.

\begin{table}[t!] 
\centering
\renewcommand{\arraystretch}{1.0}
\renewcommand{\tabcolsep}{1.6mm}
\resizebox{0.99\linewidth}{!}{
\begin{tabular}{l| c c c c}
  Method & Input size & Network & 5-shot & 1-shot \\
  \hline
  TADAM \cite{oreshkin2018tadam}               & 84  & ResNet     & 81.92 \pms{0.30} & 62.13 \pms{0.31} \\
  LEO \cite{rusu2018meta}                      & 80  & WideResNet & 81.44 \pms{0.12} & 66.33 \pms{0.09} \\
  Mean Centroid (one network)                  & 224 & ResNet     & 83.89 \pms{0.33} & 68.33 \pms{0.32} \\
  \hline
  Robust 20-\textit{dist} (ours)               & 224 & ResNet     & {\bf 85.43} \pms{0.21} & {\bf 70.44} \pms{0.32} \\
  \hline
  Robust 20 Full                               & 224 & ResNet    & 86.49 \pms{0.22} & 71.71 \pms{0.31} \\
  \hline
\end{tabular}
}
\vspace{0.3cm}
\caption{\textbf{Comparison of distilled ensembles to other methods on 1- and
    5-shot \textit{tiered}-ImageNet~\cite{ren2018meta}.} 
   To evaluate our methods we performed $5\,000$
   independent experiments on \textit{tiered}-ImageNet-test and report the average accuracy with
   $95\%$ confidence interval.  
   }
\label{tab:tiered}
\end{table}

\begin{table}[t!] 
\centering
\small\addtolength{\tabcolsep}{-10pt}
\renewcommand{\arraystretch}{1.0}
\renewcommand{\tabcolsep}{1.6mm}
\resizebox{0.99\linewidth}{!}{
\begin{tabular}{l| c c c}
  Purpose (Sign) & L2 & -cos & $\text{KL}_{\text{sim}}$ \\
  \hline
  Cooperation (+)   & 80.14 \pms{0.43} & 80.29 \pms{0.44} & 80.72 \pms{0.42}\\
  Diversity (-)  & 80.54 \pms{0.44} & 80.82 \pms{0.42} & 79.81 \pms{0.43}\\
  \hline
\end{tabular}
}
\vspace{0.3cm}
\caption{\textbf{Evaluating different relationship criteria on \textit{mini}-Imagenet 5-shot}
  The first row indicates which function was used as a relationship criteria,
  the first column indicates for which purpose the function is used and the corresponding sign. To evaluate our methods, we performed $1\,000$ independent experiments on
  CUB-test and report the average accuracy with $95\%$ confidence intervals. All ensembles
  are trained on \textit{mini}-ImageNet-train.}
\label{tab:criteria_choice}
\end{table}

When looking for diversity, the cosine similarity performs slightly better than negative L2
distance, although the accuracies are within error bars. Using negative
$\text{KL}_{\text{sim}}$ with various $\gamma$ was either not distinguishable
from independent training or was hurting the performance for larger values of $\gamma$ (not reported on the table). As for
cooperation, positive $\text{KL}_{\text{sim}}$ gives better results than L2
distance or negative cosine similarity. We believe that this behavior is due
to important difference in the way these functions compare small values in probability vectors. While negative cosine or L2
losses would penalize heavily the largest difference, $\text{KL}_{\text{sim}}$
concentrates on values that are close to 0 in one vector and are
greater in the second one.

\subsection{Performance under domain shift}\label{sec:domain_shift_exp}
Finally, we evaluate the performance of ensemble methods under domain shift.
We proceed by meta-training the models on the
\textit{mini}-ImageNet training set and evaluate the model on the CUB-test set.
The following setting was first proposed by~\cite{chen19closerfewshot} and aims
at evaluating the performance of algorithms to adapt when the difference
between training and testing distributions is large. To compare to the
results reported in the original work, we adopt their CUB test split.
Table~\ref{tab:domain_shift} compares our results to the ones listed
in~\cite{chen19closerfewshot}.
We can see that neither the full robust ensemble nor its distilled version are able to
do better than training a linear classifier on top of a frozen network. Yet, it does 
significantly better than distance-based approaches (denoted by cosine classifier in the table).
However,
if a diverse ensemble is used, it achieves the best accuracy. This is not
surprising and highlights the importance of having diverse models when
ensembling weak classifiers. 

\begin{table}[t!] 
\centering
\small\addtolength{\tabcolsep}{-10pt}
\renewcommand{\arraystretch}{1.0}
\renewcommand{\tabcolsep}{3mm}
\resizebox{0.9\linewidth}{!}{
\begin{tabular}{l| c}
  Method & \textit{mini}-ImageNet $\rightarrow$ CUB \\
  \hline
  MatchingNet \cite{vinyals2016matching} & 53.07 \pms{0.74} \\
  MAML  \cite{finn2017model}             & 51.34 \pms{0.72} \\
  ProtoNet \cite{snell2017prototypical}  & 62.02 \pms{0.70} \\
  Linear Classifier  \cite{chen19closerfewshot}   & {\bf 65.57} \pms{0.70} \\
  Cosine Classifier  \cite{chen19closerfewshot} & 62.04 \pms{0.76} \\
  Robust 20-\textit{dist}++ (ours)             & 64.23 \pms{0.58}  \\
  \hline
  \hline
  Robust 20 Full  (ours)                      & 65.04 \pms{0.57}  \\
  Diverse 20 Full (ours)                      & 66.17 \pms{0.55}  \\
\end{tabular}
}
\vspace{0.3cm}
\caption{\textbf{5-shot classification accuracy under domain shift.}
  The last two models are full ensembles and should not be
  directly compared with the rest of the table. We performed $1\,000$ independent
  experiments on \texttt{CUB-test} from~\cite{chen19closerfewshot} and report the
  average and confidence interval here. All ensembles are trained on \textit{mini}-ImageNet.}
\label{tab:domain_shift}
\end{table}

\section{Conclusions}\label{sec:ccl}
In this paper, we show that distance-based classifiers for few-shot learning
suffer from high variance, which can be significantly reduced by using an
ensemble of classifiers. Unlike traditional ensembling paradigms where diversity
of predictions is encouraged by various randomization and data augmentation techniques,
we show that encouraging the networks to cooperate during training is also important.

The overall performance of a single network obtained by distillation (with no
computational overhead at test time) leads to state-of-the-art performance for few shot 
learning, without relying on the meta-learning paradigm.  
While such a result may sound negative for meta-learning approaches, it may simply mean
that a lot of work remains to be done in this area to truly learn how to learn or to adapt.

\section*{Acknowledgment}
This work was supported by the ERC grant number 714381 (SOLARIS project), the
ERC advanced grant ALLEGRO and grants from Amazon and Intel.

{\small
\bibliographystyle{ieee_fullname}
\bibliography{biblio}
}

\clearpage
\section*{Appendix}
\addcontentsline{toc}{section}{Appendices}
\renewcommand{\thesubsection}{\Alph{subsection}}


\renewcommand\thetable{A\arabic{table}}
\renewcommand\theHtable{A\arabic{table}}
\renewcommand\thefigure{A\arabic{figure}}
\setcounter{table}{0}
\setcounter{figure}{0}

\subsection{Implementation Details}
In this section, we elaborate on training, testing and distillation details
of the proposed ensemble methods for different datasets, network architectures
and input image resolution.

\textbf{Training ResNet18 on 84x84 images on mini-ImageNet.}
For all experiments, we use ResNet18 with input image size 84x84,
train with the Adam optimizer with an
initial learning rate $3 \cdot 10^{-4}$, which is decreased by a factor 10 once during
training when no improvement in validation accuracy is observed for $p$
consecutive epochs. We use $p=20$ for training individual models, $p=30$ for
training ensembles and when distilling the model. When
distilling an ensemble into one network, $p$ is doubled. We use random crops and
color augmentation during training as well as weight decay with parameter~$\lambda=$
$5 \cdot 10^{-4}$. At training time we use random crop, color transformation and
adding random noise as data augmentation.
During the meta-testing stage, we take central crops of size $224 \times 224$
from images and feed them to the feature extractor. No other preprocessing is
used at test time. The parameters used in distillation are the same as
in Section 4.3 of the paper.

\textbf{Training WideResNet28 on 80x80 images on mini-ImageNet.}
For all experiments, we use WideResNet28 with input image size 80x80,
train with the Adam optimizer with an
initial learning rate $1 \cdot 10^{-4}$, which is decreased by a factor 10 once during
training when no improvement in validation accuracy is observed for $p$
consecutive epochs. We use $p=20$ for training individual models, $p=30$ for
training ensembles and when distilling the model. When
distilling an ensemble into one network, $p$ is doubled. We use random crops and
color augmentation during training as well as weight decay with parameter~$\lambda=$
$5 \cdot 10^{-4}$. We also set a dropout rate inside convolutional blocks to be
0.5 as described in. At training time we use random crop
and color transformation only as data augmentation.
During the meta-testing stage, we take central crops of size $80 \times 80$
from images and feed them to the feature extractor. No other preprocessing is
used at test time. The parameters used in distillation are the same as
in Section 4.3 of the paper. Here, the maximal ensemble size we evaluated
is 10 and not 20 due to memory limitations on available GPUs. Therefore, to
construct an ensemble of size 20 we merge two ensembles of size 10, that were
trained independently.

\textbf{Training ResNet18 on 224x224 images on tiered-ImageNet}
For all experiments, we use ResNet18 with input image size
224x224, train with the Adam optimizer with an initial learning rate
$3 \cdot 10^{-4}$, which is decreased by a factor 10 once during training when
no improvement in validation accuracy is observed for $p$ consecutive epochs. We
use $p=20$ for training individual models, ensembles and for distilation. We use
random crops and color augmentation during training as well as weight decay with
parameter~$\lambda=$ $1 \cdot 10^{-4}$. At training time we use random crop and
color transformation. During the meta-testing stage, we take central crops of
size $224 \times 224$ from images and feed them to the feature extractor. No
other preprocessing is used at test time. The parameters used in distillation
are the same as in Section 4.3 of the paper.

\subsection{Additional Results}
In this section we report and analyze the performance of different ensemble
types depending on their size for different network architectures and
input image resolutions.

\textbf{Few-shot Classification with ResNet18 on 224x224 images on CUB.}
The results for 1- and 5-shot classification on CUB are presented in
Table~\ref{tab:manyens_cub}. Training details and Figure summary of the results
are discussed in Experimental section of the paper.

\begin{figure*}[t!]
\begin{center}
\begin{subfigure}{.45\textwidth}
    \includegraphics[width=0.99\textwidth,trim=20 1 30 30,clip]{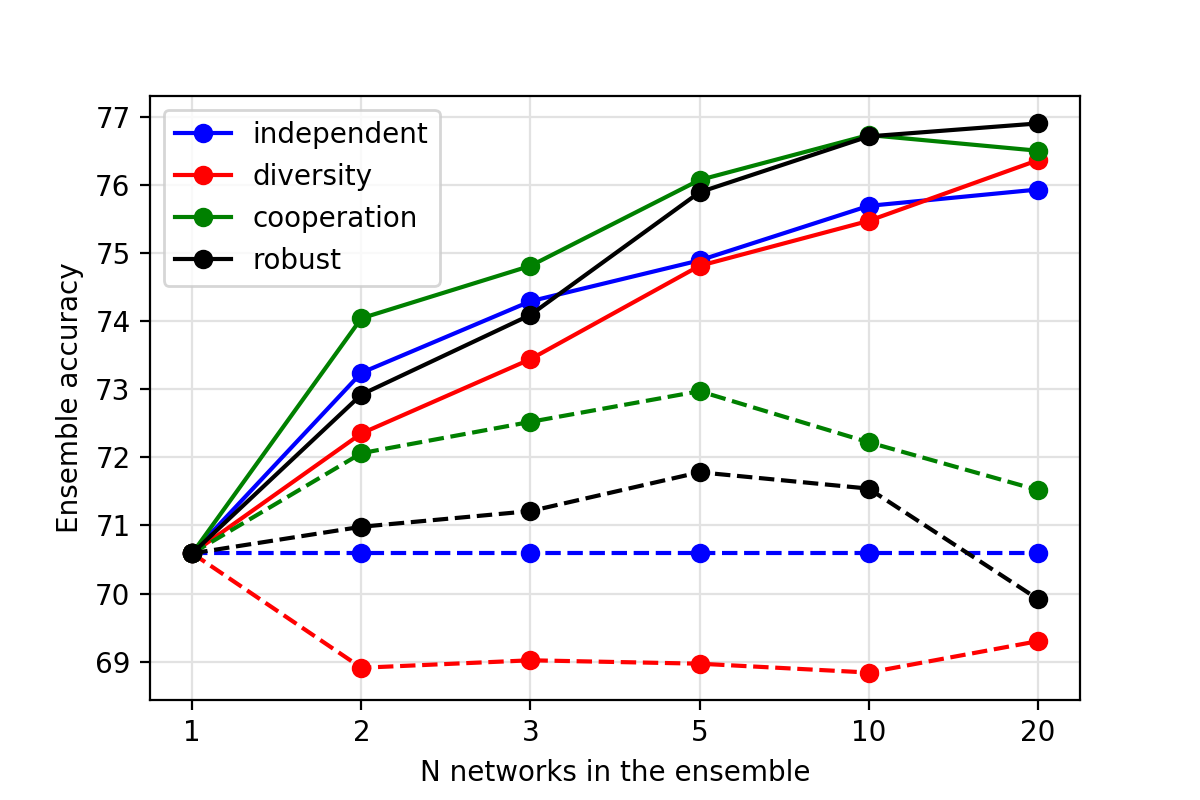}
  \caption{ResNet18 with 84x84 input}
\end{subfigure}
\begin{subfigure}{.45\textwidth}
    \includegraphics[width=0.99\textwidth,trim=20 1 30 30,clip]{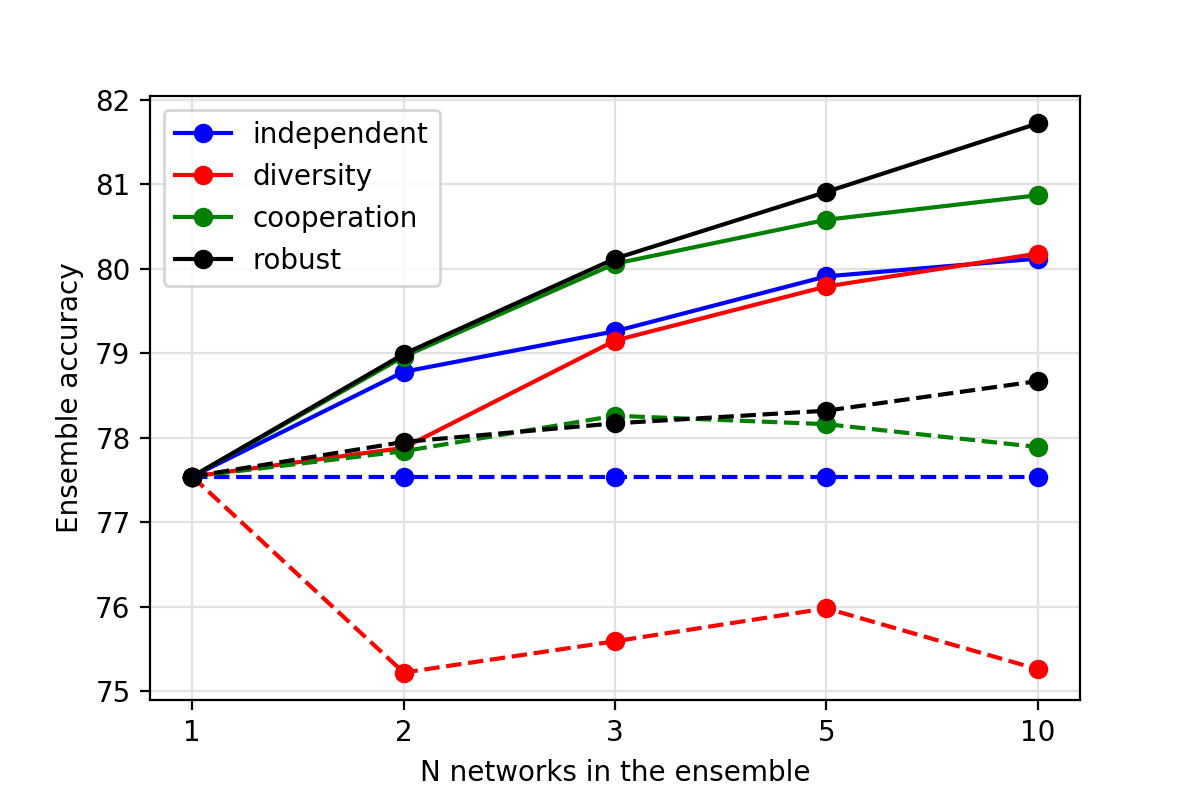}
  \caption{WideResNet28 with 80x80 input}
\end{subfigure}
\end{center}
\vspace*{-0.4cm}
\caption{\textbf{Dependency of ensemble accuracy on network architecture and
    input size for different ensemble strategies (one for each color) and various numbers
    of networks on \textit{Mini}ImageNet 5-shots classification.}
  Solid lines give the ensemble accuracy after aggregating predictions. The
  average performance of single models from the ensemble is plotted with a
  dashed line. Best viewed in color.}
\label{fig:ablation}
\vspace*{-0.3cm}
\end{figure*}

\textbf{Few-shot Classification with ResNet18 on 84x84 images on mini-ImageNet.}
The results for 1- and 5-shot classification on {\it
Mini}ImageNet are presented in Table~\ref{tab:wideresnet80} and summarized in
Figure~\ref{fig:ablation}. We can see that Cooperation training is the most
successful here for all ensemble sizes $< 20$ and other training strategies that
introduce diversity tend to perform worse. This happens because single networks
are far from overfitting the training set (as opposed to the case with 224x224
input size) and forcing diversity acts as harmful regularization. In contrary,
cooperation training enforces useful learning signal and helps ensemble members
achieve higher accuracy. Only for $n=20$ where diversity matters more, robust
ensembles perform the best.

\textbf{Few-shot Classification with WideResNet28 on 80x80 images on mini-ImageNet.}
Results for 1- and 5-shot classification on {\it Mini}ImageNet are presented in
Table~\ref{tab:resnet84} and summarized in Figure~\ref{fig:ablation}. In this
case we can see again that Diverse training does not help since the networks do
not memorize the training set. Robust ensembles outperform other training regimes
emphasizing the importance of the proposed solution that generalizes across
architectures.

\begin{table*}[t!] 
\centering
\small\addtolength{\tabcolsep}{-10pt}
\renewcommand{\arraystretch}{1.0}
\renewcommand{\tabcolsep}{1.6mm}
\resizebox{0.85\linewidth}{!}{
\begin{tabular}{l| c c c c c c}
\multicolumn{7}{c}{\bfseries 5-shot}\\ \hline
  Full Ensemble & 1 & 2 & 3 & 5 & 10 & 20 \\
  \hline
  Independent   & 79.47 \pms{0.49} & 81.34 \pms{0.46} & 82.57 \pms{0.46} & 83.16 \pms{0.45} & 83.80 \pms{0.45} & 83.95 \pms{0.46} \\
  Diversity  & 79.47 \pms{0.49} & 81.09 \pms{0.45} & 82.23 \pms{0.46} & 82.91 \pms{0.46} & 84.30 \pms{0.44} & 85.20 \pms{0.43} \\
  Cooperation     & 79.47 \pms{0.49} & 81.69 \pms{0.46} & 82.95 \pms{0.47} & 83.43 \pms{0.47} & 84.01 \pms{0.44} & 84.26 \pms{0.44} \\
  Robust  & 79.47 \pms{0.49} & 82.90 \pms{0.46} & 83.36 \pms{0.46} & 83.62 \pms{0.45} & 84.47 \pms{0.46} & 84.62 \pms{0.44} \\
  \hline
  \multicolumn{7}{l}{Distilled Ensembles}\\
  \hline
  Robust-\textit{dist}   & $-$ & 82.72 \pms{0.47} & 82.95 \pms{0.46} & 83.27 \pms{0.46} & 83.61 \pms{0.46} & 83.57 \pms{0.45} \\
  Robust-\textit{dist}++ & $-$ & 82.53 \pms{0.48} & 83.04 \pms{0.45} & 83.37 \pms{0.46} & 83.22 \pms{0.46} & 83.21 \pms{0.44} \\
  \hline
\end{tabular}
}
\resizebox{0.85\linewidth}{!}{
\begin{tabular}{l| c c c c c c}
  \multicolumn{7}{c}{\bfseries 1-shot}\\ \hline
  Ensemble type & 1 & 2 & 3 & 5 & 10 & 20 \\
  \hline
  Independent             & 64.25 \pms{0.73} & 66.60 \pms{0.72} & 67.64 \pms{0.71} & 68.07 \pms{0.70} & 68.93 \pms{0.70} & 69.64 \pms{0.69} \\

  Diversity               & 64.25 \pms{0.73} & 65.99 \pms{0.71} & 66.71 \pms{0.72} & 68.19 \pms{0.71} & 69.35 \pms{0.70} & 70.07 \pms{0.70} \\

  Cooperation             & 64.25 \pms{0.73} & 67.21 \pms{0.71} & 67.93 \pms{0.70} & 68.22 \pms{0.70} & 68.69 \pms{0.70} & 68.80 \pms{0.68} \\
  Robust                  & 64.25 \pms{0.73} & 67.33 \pms{0.71} & 68.01 \pms{0.72} & 68.53 \pms{0.70} & 68.59 \pms{0.70} & 69.47 \pms{0.69} \\
  \hline
  \multicolumn{7}{l}{Distilled Ensembles}\\
  \hline
  Robust-\textit{dist}    & $-$ & 67.47 \pms{0.71} & 67.29 \pms{0.72} & 68.09 \pms{0.70} & 68.71 \pms{0.71} & 68.77 \pms{0.71} \\
  Robust-\textit{dist}++  & $-$ & 67.01 \pms{0.74} & 67.62 \pms{0.72} & 68.68 \pms{0.71} & 68.38 \pms{0.70} & 68.68 \pms{0.69} \\
  \hline
\end{tabular}
}

\vspace{0.3cm}
\caption{\textbf{Few-shot classification accuracy on CUB.}
  The first column gives the type of ensemble and the top row indicates the number
  of networks in an ensemble. Here, \textit{dist} means that an ensemble was
  distilled into a single network, and '++' indicates that extra unannotated
  images were used for distillation. We performed 1000 independent experiments on
  CUB-test and report the average with 95\% confidence interval. All
  networks are trained on CUB-train set.}
\label{tab:manyens_cub}
\end{table*}

\begin{table*}[t!] 
\centering
\small\addtolength{\tabcolsep}{-10pt}
\renewcommand{\arraystretch}{1.0}
\renewcommand{\tabcolsep}{1.6mm}
\resizebox{0.85\linewidth}{!}{
\begin{tabular}{l| c c c c c c}
\multicolumn{7}{c}{\bfseries 5-shot}\\ \hline
  Ensemble type & 1 & 2 & 3 & 5 & 10 & 20\\
  \hline
  Independent & 70.59 \pms{0.51} & 73.24 \pms{0.49} & 74.29 \pms{0.48} & 74.89 \pms{0.47}&  75.69 \pms{0.47} & 75.93 \pms{0.47} \\

  Diversity   & 70.59 \pms{0.51} & 72.35 \pms{0.47} & 73.44 \pms{0.49} & 74.81 \pms{0.48}&  75.47 \pms{0.48} & 76.36 \pms{0.47} \\

  Cooperation & 70.59 \pms{0.51} & 74.04 \pms{0.47} & 74.81 \pms{0.47} & 76.37 \pms{0.48}&  76.73 \pms{0.48} & 76.50 \pms{0.47} \\
  \hline                                         
  Robust      & 70.59 \pms{0.51} & 72.92 \pms{0.50} & 73.09 \pms{0.43} & 75.69 \pms{0.42}&  76.71 \pms{0.47} & 76.90 \pms{0.48} \\
  \hline
  \multicolumn{7}{l}{Distilled Ensembles}\\
  \hline
  Robust-\textit{dist}     & $-$ & 73.04 \pms{0.50} & 73.58 \pms{0.49} & 74.35 \pms{0.48}&  74.69 \pms{0.49} & 75.24 \pms{0.49} \\
  Robust-\textit{dist}++   & $-$ & 73.50 \pms{0.49} & 74.17 \pms{0.49} & 74.84 \pms{0.49}&  75.12 \pms{0.44} & 75.62 \pms{0.48} \\
  \hline
\end{tabular}
}
\vspace{0.4cm}
\resizebox{0.85\linewidth}{!}{
\begin{tabular}{l| c c c c c c}
  \multicolumn{7}{c}{\bfseries 1-shot}\\ \hline
  Ensemble type & 1 & 2 & 3 & 5 & 10 & 20 \\
  \hline
  Independent & 53.31 \pms{0.64} & 55.72 \pms{0.60} & 56.85 \pms{0.64} & 57.90 \pms{0.63} & 58.21 \pms{0.63} & 58.56 \pms{0.61} \\

  Diversity   & 53.31 \pms{0.64} & 54.61 \pms{0.62} & 55.90 \pms{0.62} & 57.06 \pms{0.63} & 57.49 \pms{0.62} & 58.93 \pms{0.64} \\

  Cooperation & 53.31 \pms{0.64} & 55.80 \pms{0.64} & 57.13 \pms{0.63} & 58.18 \pms{0.64} & 58.63 \pms{0.63} & 58.73 \pms{0.62} \\
  Robust      & 53.31 \pms{0.64} & 55.95 \pms{0.62} & 56.27 \pms{0.64} & 58.51 \pms{0.65} & 59.38 \pms{0.65} & 59.48 \pms{0.65} \\
  \hline
  \multicolumn{7}{l}{Distilled Ensembles}\\
  \hline
  Robust-\textit{dist}     & $-$ & 56.84 \pms{0.64} & 56.58 \pms{0.65} & 57.13 \pms{0.63} & 57.41 \pms{0.65} & 58.11 \pms{0.64} \\
  Robust-\textit{dist} ++  & $-$ & 56.53 \pms{0.62} & 57.03 \pms{0.64} & 57.48 \pms{0.65} & 58.05 \pms{0.63} & 58.67 \pms{0.65} \\
  \hline
\end{tabular}
}
\vspace{0.1cm}
\caption{\textbf{Few-shot classification accuracy on \textit{Mini}ImageNet,
    using ResNet18 and 84x84 image size.}
  The first column gives the strategy, the top row indicates the number~$N$
  of networks in an ensemble. Here, \textit{dist} means that an ensemble was
  distilled into a single network, and '++' indicates that extra unannotated
  images were used for distillation. We performed 1\,000 independent experiments on
  {\it Mini}ImageNet-test and report the average with 95\% confidence interval.
All networks are trained on {\it Mini}ImageNet-train set.}
\label{tab:resnet84}
\end{table*}

\begin{table*}[t] 
\centering
\small\addtolength{\tabcolsep}{-10pt}
\renewcommand{\arraystretch}{1.0}
\renewcommand{\tabcolsep}{1.6mm}
\resizebox{0.85\linewidth}{!}{
\begin{tabular}{l| c c c c c}
\multicolumn{6}{c}{\bfseries 5-shot}\\ \hline
  Ensemble type & 1 & 2 & 3 & 5 & 10 \\
  \hline
  Independent & 77.54 \pms{0.45} & 78.78 \pms{0.45} & 79.26 \pms{0.43} & 79.91 \pms{0.44} & 80.12 \pms{0.43} \\

  Diversity   & 77.54 \pms{0.45} & 77.88 \pms{0.45} & 79.15 \pms{0.44} & 79.79 \pms{0.44} & 80.18 \pms{0.44} \\

  Cooperation & 77.54 \pms{0.45} & 78.96 \pms{0.46} & 80.06 \pms{0.44} & 80.58 \pms{0.45} & 80.87 \pms{0.43} \\
  \hline                                        
  Robust      & 77.54 \pms{0.45} & 78.99 \pms{0.45} & 80.12 \pms{0.43} & 80.91 \pms{0.43} & 81.72 \pms{0.44} \\
  \hline
  \multicolumn{6}{l}{Distilled Ensembles}\\
  \hline
  Robust-\textit{dist}     & $-$ & 79.44 \pms{0.44} & 79.84 \pms{0.44} & 80.01 \pms{0.42} & 80.85 \pms{0.43} \\
  Robust-\textit{dist}++  & $-$ & 79.16 \pms{0.46} & 80.00 \pms{0.44} & 80.25 \pms{0.42}&  81.11 \pms{0.43} \\
  \hline
\end{tabular}
}
\vspace{0.4cm}
\resizebox{0.85\linewidth}{!}{
\begin{tabular}{l| c c c c c}
  \multicolumn{6}{c}{\bfseries 1-shot}\\ \hline
  Ensemble type & 1 & 2 & 3 & 5 & 10 \\
  \hline
  Independent & 59.02 \pms{0.63} & 60.07 \pms{0.62} & 60.58 \pms{0.61} & 61.24 \pms{0.63} & 62.05 \pms{0.61} \\

  Diversity   & 59.02 \pms{0.63} & 58.87 \pms{0.62} & 60.63 \pms{0.61} & 61.30 \pms{0.62} & 62.28 \pms{0.61} \\

  Cooperation & 59.02 \pms{0.63} & 60.22 \pms{0.62} & 61.03 \pms{0.61} & 62.07 \pms{0.61} & 62.42 \pms{0.61} \\
  Robust      & 59.02 \pms{0.63} & 60.92 \pms{0.62} & 62.03 \pms{0.62} & 62.78 \pms{0.61} & 63.39 \pms{0.62} \\
  \hline
  \multicolumn{6}{l}{Distilled Ensembles}\\
  \hline
  Robust-\textit{dist}     & $-$ & 61.07 \pms{0.62} & 61.57 \pms{0.61} & 62.24 \pms{0.61} & 62.80 \pms{0.62} \\
  Robust-\textit{dist} ++  & $-$ & 61.37 \pms{0.62} & 62.01 \pms{0.60} & 62.45 \pms{0.62} & 63.25 \pms{0.62} \\
  \hline
\end{tabular}
}
\vspace{0.1cm}
\caption{\textbf{Few-shot classification accuracy on \textit{Mini}ImageNet,
    using WideResNet28 and 80x80 image size.}
  The first column gives the strategy, the top row indicates the number~$N$
  of networks in an ensemble. Here, \textit{dist} means that an ensemble was
  distilled into a single network, and '++' indicates that extra unannotated
  images were used for distillation. We performed 1\,000 independent experiments on
  {\it Mini}ImageNet-test and report the average with 95\% confidence interval.
All networks are trained on {\it Mini}ImageNet-train set.}
\label{tab:wideresnet80}
\end{table*}

\end{document}